\documentclass[10pt]{article} % For LaTeX2e
\usepackage[preprint]{tmlr}

\usepackage{amsmath,amsfonts,bm}

\def\eqref#1{equation~\ref{#1}}
\def\1{\bm{1}}

\DeclareMathAlphabet{\mathsfit}{\encodingdefault}{\sfdefault}{m}{sl}
\SetMathAlphabet{\mathsfit}{bold}{\encodingdefault}{\sfdefault}{bx}{n}

\usepackage{url}
\usepackage{comment}
\usepackage{algorithm}
\usepackage{algpseudocode}
\usepackage{bbm}
\usepackage{colortbl}
\usepackage{tikz}
\usepackage{array}
\newcolumntype{H}{>{\setbox0=\hbox\bgroup}c<{\egroup}@{}}
\newcommand*\circled[1]{\tikz[baseline=(char.base)]{
            \node[shape=circle,draw,inner sep=1pt] (char) {#1};}}
\newcommand{\cA}{\circled{1}~}
\newcommand{\cB}{\circled{2}~}
\newcommand{\cC}{\circled{3}~}

\algnewcommand{\LineComment}[1]{\State \(\triangleright\) #1}
\usepackage[textsize=tiny,textwidth=15mm]{todonotes}

\title{Adaptive Training Distributions\\ with Scalable Online Bilevel Optimization}

\author{\name David Grangier, Pierre Ablin, Awni Hannun\email \{grangier, p\_ablin, awni\}@apple.com \\
      \addr Apple \\
}

\def\month{MM}  % Insert correct month for camera-ready version
\def\year{YYYY} % Insert correct year for camera-ready version
\def\openreview{\url{https://openreview.net/forum?id=XXXX}} % Insert correct link to OpenReview for camera-ready version

\begin{document}

\maketitle

\begin{abstract}
Large neural networks pretrained on web-scale corpora are central to modern machine learning. In this paradigm, the distribution of the large, heterogeneous pretraining data rarely matches that of the application domain. This work considers modifying the pretraining distribution in the case where one has a small sample of data reflecting the targeted test conditions. We propose an algorithm motivated by a recent formulation of this setting as an online, bilevel optimization problem. With scalability in mind, our algorithm prioritizes computing gradients at training points which are likely to most improve the loss on the targeted distribution. Empirically, we show that in some cases this approach is beneficial over existing strategies from the domain adaptation literature but may not succeed in other cases. We propose a simple test to evaluate when our approach can be expected to work well and point towards further research to address current limitations.
\end{abstract}

\section{Introduction}

Large models pretrained on massive, heterogeneous datasets 
have impacted various application domains~\citep{bommasani2021foundation}, 
including natural language processing~\citep{devlin2019bert}, 
computer vision~\citep{mahajan2018pretraining}, and audio 
processing~\citep{schneider2019wav2vec}. 
These models are typically trained on two different distributions: a {\it generic 
distribution} for pretraining and a {\it specific distribution} for fine tuning. Only 
the specific distribution matches the test conditions while the generic 
distribution offers an abundant source of data with some similarities to the specific data.
This novel paradigm builds upon earlier work in multitask 
learning~\citep{caruana1997multitask}, transfer 
learning~\citep{bennett2003inductive}, and
domain adaptation~\citep{moore-lewis-2010-intelligent}. For all of these
methods, the accuracy of a model on the specific task heavily
depends on selecting an appropriate distribution over the generic auxiliary
tasks and data.

This work proposes a scalable online strategy for data selection along with a comprehensive and realistic empirical study. We build upon a bilevel formulation of the generic re-weighting problem which allows for gradient-based optimization \citep{franceschi2018bilevel}.

Our contributions are several. First, we unify several gradient-based data selection methods into a common framework in which their similarities and distinctions are more easily understood. Second, we introduce a scalable, online algorithm. This algorithm can train a large model while updating an inexpensive 
auxiliary data selection model which tracks the distribution required to make fast progress on the targeted task. Our algorithm leverages the asymmetry in computational cost between the selection model and the large model by filtering examples on the fly, ensuring that the majority of examples are not examined by the large model.

Third, we perform a comprehensive and realistic empirical comparison of data selection strategies. We compare several alternative strategies across different tasks and modalities including large scale language modeling, machine translation, and image classification. Finally, we propose a simple metric based on gradient alignment that correlates with the success and failure of gradient-based data selection methods.

\section{Related Work}

Prior work has proposed automatic methods to adjust the generic 
training distribution in order to improve model generalization on
the specific task. The domain adaptation literature has explored
variants of importance sampling, which uses importance weights to 
emphasize or select some generic examples.
These weights have been determined via 
domain classifiers~\citep{aharoni-goldberg-2020-unsupervised,gururangan2020dontstoppretraining2020},
via gradient alignment and 
fine-tuning~\citep{wang2018dataselection,grangier2022tradeoffs},
or via the estimation of the label distribution~\citep{ngiam2018domain}.
Related to domain adaptation, the removal of label noise in the
generic distribution has received attention with methods based 
on influence 
functions~\citep{koh2017influence,pruthi2020estimating,schioppa2022scaling}, 
data models~\citep{ilyas2022datamodels,jain2022data}, 
and data Shapley values~\citep{ghorbani2019data,karlas2022arxiv}.

As an alternative to static weighting, the literature
also explored dynamic weighting where the distribution
over generic examples is adapted during training. The two primary strategies are reinforcement learning and direct optimization.
Reinforcement learning does not assume that the specific task
loss can be differentiated with respect to the weighting
parameters. Instead, a parameterized model of the generic distribution 
is adjusted through reinforcement learning: the current model proposes
generic distributions, and their reward is measured as 
the specific loss after a few steps of generic training 
over a proposal distribution~\citep{%
kumar2019reinforcement,yoon2020data_valuation,zhu2020learning_to_transfer}.
On the other hand, direct optimization assumes 
a differentiable functional dependency between the weighting parameters 
and the specific training loss. This dependency can be derived through
meta learning by unfolding the 
generic
update~\citep{ren2018reweight,hu2019learning,shu19metaweight,zhang_pfister21}: 
one gradient update step minimizing the weighted generic
loss depends on the weighting parameters. The impact of this update can 
be evaluated by computing the post-update specific loss which can then be 
differentiated with respect to the weighting parameters. As an alternative to
update unfolding, a bilevel formulation of the 
reweighting problem also allows for direct optimization \citep{franceschi2018bilevel}.
Our work builds upon this bilevel formulation.

Other research areas intersect with generic sample reweighting.
Prior work considered learning a distribution over training data 
augmentations~\citep{ho2019population,lim2019fast,zoph2020learning}.
Curriculum learning has also been introduced to visit successive training 
distributions based on training instance difficulty~\citep{bengio2009curriculum,kumar2010self_paced,jiang2018mentornet,saxena:data_parameters}.
Multi-task learning research has considered gradient projection 
to minimize negative interactions between tasks~\citep{yu2020gradient,dery2020auxiliary,liu2021conflict}.
Importance sampling for accelerated stochastic training~\citep{zhao2015stochastic,katharopoulos2018importance} is also relevant.

\section{Problem Setting}

Classical machine learning assumes that the model is trained on data drawn from the distribution 
from which the test data will also be sampled from~\citep{vapnik1999nature}. Our setting is different 
and belongs to the field of transfer learning~\citep{caruana1993multitask, thrun1998learning}. In our setting 
we are given two training sets, a large 
generic training set $\mathcal{D}_{\text{generic}}$ and small specific training set 
$\mathcal{D}_{\text{specific}}$. Only the latter set is representative of the test conditions. The
large generic set can be leveraged as it might contain information related to the targeted specific.
Its large scale allows more reliable statistical estimation and allows training higher capacity models.

Domain adaptation~\citep{farahani2021brief}, multi-task learning~\citep{caruana1993multitask}, fine-tuning~\citep{denevi2018learning} are transfer learning setups 
related to our setting, see Appendix~\ref{sec:transfer_learning} for a discussion on the relevant terminology. It is important to note that in our setting the targeted task is known at the beginning of learning (which is not a 
necessity for fine tuning) and that only the specific task accuracy matters (unlike multi-task learning which
might also target high generic accuracy). In this work, we present our work while targeting a single 
specific task, with the same loss as the generic loss. Our derivations can be easily extended to the 
case where the specific and generic task have different loss functions. It is also simple to write
derivations for the case where the specific loss is a mixture over multiple targeted specific tasks.
We leave this extensions for future work.

\section{Methods}
\label{sec:methods}

We aim to identify the parameters $\theta$ of a model that
achieves good generalization performance (held-out likelihood) over the
specific distribution. For that purpose, we are given a large generic training
set $\mathcal{D}_{\text{generic}}$ and small specific training set 
$\mathcal{D}_{\text{specific}}$. We propose to formulate the generic training problem as the
minimization of the weighted loss,
$$
\mathcal{L}_{\text{generic}}(\theta, \alpha)=\sum_{x\in\mathcal{D}_{\text{generic}}}w(x; \alpha)\ell(x; \theta)
$$
where $w(x; \alpha)$ denotes a smaller, secondary {\it weighting neural network} which defines a distribution over $\mathcal{D}_{\text{generic}}$, i.e. 
$\forall x, w(x; \alpha) > 0$ and $\sum_{x\in\mathcal{D}_{\text{generic}}}w(x; \alpha) = 1$.
We denote the solution to generic training problem as
\begin{equation}
    \label{eq:inner_problem}
\theta^*(\alpha)\in \arg\min_\theta\mathcal{L}_{\text{generic}}(\theta, \alpha)
\end{equation}
Our goal is to find the parameter of the weighting network such that the loss on the
{\it specific} training set is minimal, i.e. minimizing,
\begin{equation}
    \label{eq:outer_problem}
    \mathcal{L}_{\text{specific}}(\theta^*(\alpha)):=\sum_{x'\in\mathcal{D}_{\text{specific}}}\ell(x'; \theta^*(\alpha)).
\end{equation}
with respect to $\alpha$.

\subsection{Data Selection as a Bilevel Optimization Problem}

Our notations make clear that finding the optimal weighting network can be cast as a bilevel optimization problem: 
with a fixed weighting network, the optimal parameters for the main model are found by minimizing the weighted loss 
over the generic dataset, $\mathcal{L}_{\text{generic}}$ (Equation~\ref{eq:inner_problem}). The optimal main model parameters $\theta^*$ depends explicitly on the weighting network parameters $\alpha$; indeed, changing
$\alpha$ changes the optimization problem in~Equation~\ref{eq:inner_problem} and its solution. The selection of $\alpha$ is driven by the specific set loss, $\mathcal{L}_{\text{specific}}$(Equation~\ref{eq:outer_problem}).

Equation~\ref{eq:inner_problem} and Equation~\ref{eq:outer_problem} form a \emph{bilevel optimization problem}~\citep{franceschi2018bilevel}: the outer problem (Equation~\ref{eq:outer_problem}) depends implicitly on $\alpha$ through the solution to the inner problem (\ref{eq:inner_problem}).
One of the strengths of such a bilevel formulation is that the weighting network must adapt to the main model: the question is to learn a weighting network such that the main model trained with that network leads to good specific performance.
This has the potential to go beyond a simple model-agnostic scheme that would, for instance, build $w(x)$ based on the similarity between $x$ and the specific set.  
While a large body of the literature is devoted to solving bilevel problems where the inner problem (Equation~\ref{eq:inner_problem} is convex in $\theta$~\citep{ghadimi2018approximation,arbel2021amortized}, in our case, Equation~\ref{eq:inner_problem} corresponds to the training problem of a neural network which is non-convex. 
This leads to several difficulties:
\begin{itemize}
    \item The $\arg\min$ in Equation~\ref{eq:inner_problem} is not a single element since there are multiple minimizers. 
    Therefore, the function $\theta^*(\alpha)$ is not properly defined.
    \item In order to use gradient-based methods to find the optimal $\alpha$, we have to compute the approximate Jacobian of $\theta^*(\alpha)$. 
    This is usually done using the implicit function theorem, which only applies when the loss function in~\eqref{eq:inner_problem} is locally convex and
    such property is hard to check in practice.
\end{itemize}
Furthermore, we want a method with a computational cost similar to the standard training of the main model. In other words, we have enough budget to solve Equation~\ref{eq:inner_problem} only once: learning $\alpha$ and $\theta$ must be carried out synchronously.
This has an important consequence: the bilevel methods that we study update $\alpha$ based on the current main model state $\theta$ and not on the optimal solution $\theta^*(\alpha)$.
Hence, this is a slight deviation from the bilevel formalism. This also means that the weighting network adapts to the current state of the main model and, ideally, tries to up-weight generic data that is useful \emph{at the current state of learning}.
%
% \subsection{Bilevel Methods for Data Selection}
%
We explore online algorithms to solve the bilevel problem when the main model is large.
These algorithms alternate $\theta$ and $\alpha$ updates and leverage the asymmetry in computation cost 
between evaluating the large main model and the small auxiliary weighting network.
\subsection{Updating the main model}
\label{subsec:big_batch_trick}
 To update the main model, we fix $\alpha$ and do a step to minimize Equation~\ref{eq:inner_problem}.
A first, natural idea would be to take a mini-batch of generic data $B_{\text{generic}}$ of size $b$, compute the corresponding gradient $g = \frac1b\sum_{x\in B_{\text{generic}}}w(x; \alpha) \nabla_\theta \ell(x; \theta)$ and then use it to update $\theta$, either implementing SGD by doing $\theta \leftarrow \theta - \eta \times g$ with $\eta>0$ a learning rate, or by using it into a more involved optimizer like Adam.
However, the computation of $g$ with the previous equation can be wasteful when a significant fraction of 
the examples of $B_{\text{generic}}$ are assigned small weights $w(x; \alpha)$. These examples do not contribute much to $g$ while still requiring the expensive computation of their gradient $\nabla_\theta \ell(x; \theta)$.

To accelerate the optimization of $\theta$, we leverage the asymmetry between the cost of evaluating the weighting network and the main model: computing $w(x;\alpha)$ only requires inference of a small network while computing $\nabla \ell(x; \theta)$ requires inference \emph{and} back-propagation through a large network.
We start by sampling a large batch $B^{\text{big}}_{\text{generic}}$ from the generic dataset and compute $w(x;\alpha)$ for each $x$ in $B^{\text{big}}_{\text{generic}}$. From there we can take a smaller batch $B^{\text{small}}_{\text{generic}}$ from
$B^{\text{big}}_{\text{generic}}$, either by sampling from the distribution defined by $w(x;\alpha)$ or by taking the examples with the highest $w(x;\alpha)$. The first option is an unbiased solution corresponding to importance sampling, while the second option is biased but observed to work better in practice.
In both cases, we compute the gradient to update $\theta$ with uniform weights, using $g = \frac1b\sum_{x\in B^{\text{small}}_{\text{generic}}}\nabla_\theta \ell(x; \theta)$.

\subsection{Updating the weighting model}
With scalability in mind, 
we only consider \emph{stochastic} methods, i.e., that update the weighting network parameters $\alpha$ using only a mini-batch of specific data $B_{\text{specific}}$ and a mini-batch of generic data $B_{\text{generic}}$.
We consider three alternatives to update the weighting model. 

Before describing alternative methods to update $\alpha$, we summarize our approach in Algorithm~\ref{alg:online_bilevel_selection}.
We denote ${\rm  sample}(D, n)$ the set resulting from sampling $n$ times uniformly from a set $D$. We denote ${\rm filter}(D, \alpha,  n)$
the result from either (a) sampling $n$ times from $D$ i.i.d relying on the distribution induced by the weighting model at $\alpha$, or (b) selecting the top-$n$ highest weighted examples from $D$. The batch sizes $b_{\text{small}}, b_{\text{large}}$ are hyper-parameters selected through validation.
\begin{center}
\begin{minipage}{.7\linewidth}
\begin{algorithm}[H]
\caption{Scalable, Online Bilevel Data Selection}\label{alg:online_bilevel_selection}
\begin{algorithmic}
\Require $\mathcal{D}_{\text{generic}}, \mathcal{D}_{\text{specific}}, b_{\text{small}}, b_{\text{large}}$ \Comment{Training datasets, batch sizes.}
\State $\theta_0 \gets {\rm main\_model\_initializer()}$
\State $\alpha_0 \gets {\rm weight\_model\_initializer()}$
\For{$t = 1, \ldots, T$}
\LineComment{Sample generic and specific batch.}
\State $B_{\text{generic}} \gets {\rm  sample}(\mathcal{D}_{\text{generic}}, b_{\text{large})}$
\State $B_{\text{specific}} \gets {\rm  sample}(\mathcal{D}_{\text{specific}},  b_{\text{small}})$
\\\LineComment{Sample generic sub-batches.}
\State $B_{\text{filtered}} \gets {\rm filter}(B_{\text{generic}}, \alpha_{t-1},  b_{\text{small}})$
\State $B'_{\text{generic}} \gets {\rm sample}(B_{\text{generic}},  b_{\text{small}})$

\\\LineComment{Inner and outer updates.}
\State $\theta_t \gets {\rm update\_main\_model}(B_{\text{filtered}}, \theta_{t-1})$
\State $\alpha_t \gets {\rm update\_weight\_model}(B'_{\text{generic}}, B_{\text{specific}}, \theta_{t}, \alpha_{t-1})$
\EndFor
\State \Return $\theta_T$ \Comment{Trained main model.}
\end{algorithmic}
\end{algorithm}
\end{minipage}
\end{center}

\subsubsection{One gradient step unrolling - differentiable data selection (DDS)}
This method is similar to~\citep{wang2020optimizing}, and updates the weighting network by doing a descent step on the loss
\begin{equation}
    \label{eq:loss_dds}
    \mathcal{L}(\alpha) = \sum_{x'\in B_{\text{specific}}}\ell'(x'; u(\theta, \alpha)) \text{ with }u(\theta; \alpha) = \theta - \rho \times \sum_{x\in B_{\text{generic}}}w(x; \alpha)\nabla_\theta \ell(x; \theta),
\end{equation}
which corresponds to the value of the specific loss on the mini-batch $B_{\text{specific}}$ after a gradient descent step for $\theta$ on the generic mini-batch $B_{\text{generic}}$ using the current weights.
The idea behind this method is that $u(\theta, \alpha)$ is a reasonable approximation to $\theta^*(\alpha)$.
This method requires backpropagating through a gradient descent step, which requires only a little overhead compared to a standard gradient computation.
In the limit where the step size $\rho$ in the gradient update $u(\theta, \alpha)$ goes to $0$, we see that $\mathcal{L}(\alpha)\simeq \rho\langle g_{\text{specific}}, g_{\text{generic}}\rangle$, with $g_{\text{specific}} = \sum_{x'\in B_{\text{specific}}}\nabla\ell'(x'; \theta)$ and $g_{\text{generic}}=\sum_{x\in B_{\text{generic}}}w(x; \alpha)\nabla\ell(x, \theta)$.
Hence, the loss $\mathcal{L}$ approximately measures the alignement between specific and generic gradients.
Taking derivatives gives $\nabla\mathcal{L}(\alpha)\simeq \rho\sum_{x\in B_{\text{generic}}}\langle g_{\text{specific}}, \nabla\ell(x, \theta)\rangle\nabla w(x; \alpha)$.

\subsubsection{Stochastic Bilevel Algorithm (SOBA)}
We also implement the SOBA method of~\citep{dagreou2022bilevel}, which is a scalable method to solve the bilevel problem, developed in a setting where the inner function (Equation~\ref{eq:inner_problem}) is convex. 
This algorithm approximates a gradient descent on $h(\alpha) = \mathcal{L}_{\text{specific}}(\theta^*(\alpha))$.
The chain rule gives $\nabla h(\alpha) = \frac{\partial \theta^*}{\partial\alpha}\nabla\mathcal{L}_{\text{specific}}(\theta^*(\alpha))$. The optimum $\theta^*(\alpha)$ satisfies the first order condition $\nabla_\theta \mathcal{L}_{\text{generic}}(\theta^*(\alpha), \alpha) = 0$. 
Under the assumption that the Hessian $\nabla^2_{\theta\theta}\mathcal{L}_{\text{generic}}(\theta^*(\alpha), \alpha)$ is invertible, the implicit function theorem applied to the previous equation gives $\frac{\partial \theta^*}{\partial\alpha} = - \nabla^2_{\alpha\theta}\mathcal{L}_{\text{generic}}(\theta^*(\alpha), \alpha)\left[\nabla^2_{\theta\theta}\mathcal{L}_{\text{generic}}(\theta^*(\alpha), \alpha)\right]^{-1}$, which overall yields $\nabla h(\alpha) = -\nabla^2_{\alpha\theta}\mathcal{L}_{\text{generic}}(\theta^*(\alpha), \alpha)\left[\nabla^2_{\theta\theta}\mathcal{L}_{\text{generic}}(\theta^*(\alpha), \alpha)\right]^{-1}\nabla\mathcal{L}_{\text{specific}}(\theta^*(\alpha))$. 
SOBA approximates this quantity in two ways: first, $\theta^*(\alpha)$ is replaced by the current iterate $\theta$ in the above gradient. 
Second, in addition to $\theta$ and $\alpha$, SOBA has an additional variable $v$ of the same size as $\theta$ that keeps track of the quantity $-\left[\nabla^2_{\theta\theta}\mathcal{L}_{\text{generic}}(\theta, \alpha)\right]^{-1}\nabla_{\theta}\mathcal{L}_{\text{specific}}(\theta)$. 
This is done using the stochastic iterations $v\leftarrow v - \eta \times dv$ with $dv 
 = \sum_{x\in B_{\text{generic}}}w(x; \alpha)\nabla^2\ell(x;\theta)v + \sum_{x'\in B_{\text{specific}}}\nabla \ell'(x';\theta)$. 
 The first part in $dv$ is a Hessian-vector product that can be computed efficiently at a cost similar to that of a gradient~\citep{pearlmutter1994fast}. 
 Then, the parameters $\alpha$ are moved in the direction $d\alpha = \sum_{x\in B_{\text{generic}}}\langle \nabla\ell(x;\theta), v\rangle \nabla w(x;\alpha)$, which is a stochastic approximation of $\nabla^2_{\alpha\theta}\mathcal{L}_{\text{generic}}(\theta, \alpha)v$, which is itself an approximation of $\nabla h(\alpha)$.

\subsubsection{Aligned NOrmalized GRADient (Anograd)}

We derive Anograd (Aligned NOrmalized GRADient) as a variant of DDS which relies on steepest descent~\citep{boyd2004convex}. We apply the steepest descent algorithm to the specific loss $\theta \to \mathcal{L}_{\text{specific}}(\theta)$. 
We recall that the steepest normalised descent direction according to the Euclidean norm $\| \cdot \|$ for the specific 
loss is 
\begin{equation}
\Delta \theta_{\rm nsd} = \arg\min_v \{ v^\top \nabla_\theta \mathcal{L}_{\text{specific}}(\theta) : \| v \|=1\}
\label{eq:nsd}
\end{equation}
This direction aligns with the opposite gradient when $v$ is not further constrained. In our case,
$\theta$ updates should correspond to gradient descent updates on the weighted generic loss. We 
therefore constraints $\theta$ updates 
to decompose as an afine combination of individual generic example gradients, i.e.
$
\sum_{i=1}^{n_{\text{g}}} a_i \nabla\ell(x^{\text{g}}_i, \theta)
$
where ${\cal D}_{\text{generic}}$ is denoted as $\{x^{\text{g}}_i\}_{i=1}^{n_{\text{g}}}$ and $a_i > 0, \forall i $.
Therefore, we need 
to solve Equation~\ref{eq:nsd} with the constraint $v \in {\cal V}$, 
with
$
{\cal V}=
\left\{ \sum_{i=1}^{n_{\text{g}}} a_i \nabla\ell(x^{\text{g}}_i, \theta), \forall a_i \ge 0\right\}.
$
This amounts to solving
$$
\min_a
\left(
\frac{\sum_{i=1}^{n_{\text{g}}} a_i \nabla \ell(x^{\text{g}}_i, \theta)}{\|\sum_{i=1}^{n_{\text{g}}} a_i \nabla \ell(x^{\text{g}}_i, \theta))\|}
\right)^\top
\nabla_\theta \mathcal{L}_{\text{specific}}(\theta)
$$
which itself is equivalent to solve
$
\min_a ~
\text{cosine}\left(
\sum_{i=1}^{n_{\text{g}}} a_i \nabla \ell(x^{\text{g}}_i, \theta),~
\nabla_\theta \mathcal{L}_{\text{specific}}(\theta)\right)
$
We now parameterize $a$ as the output of the weighting network
and introduce the loss,
$$
{\cal L}_{\text{anograd}}(\theta, \alpha)
=
\text{cosine}\left(\nabla_\theta \mathcal{L}_{\text{generic}}(\theta, \alpha),~
\nabla_\theta \mathcal{L}_{\text{specific}}(\theta)\right).
$$
The anograd method performs gradient descent on that loss to update $\alpha$. Like for 
DDS and Soba, we perform a single step before updating $\theta$.
For scalability we rely on stochastic (batch) estimates for both
both terms in the cosine. Compared to DDS, the normalization in anograd reduces
the benefit of up-weighting generic examples with high gradient norm.

\section{Experiments \& Results}
\label{sec:experiments}

Our experiments focus on three application domains: language modeling, machine translation and image classification. 
Before introducing our experimental setup and discussing our results on each domain, we describe the baselines we considered.
% In all three cases we compare our results with contrastive 
% data selection, CDS~\citep{van-der-wees-etal-2017-dynamic,wang2018dataselection}, domain classifier selection~\citep{gururangan2020dontstoppretraining2020}, meta-weight net~\citep{shu19metaweight}).
% and learning to re-weight, LTR~\citep{ren2018reweight}.

\subsection{Evaluated Alternative Methods}

For our empirical comparison, we first consider two common, simple methods which do not rely on data selection.
We call {\it baseline} pretraining on the generic training set followed by fine tuning on the specific set.
We call {\it mixing} pretraining on a mix of generic and specific data. Each training batch contains a fixed 
fraction of specific data. This fraction is selected by validation.

Among data selection alternatives, we first consider {\it contrastive data selection}, CDS~\citep{moore-lewis-2010-intelligent,van-der-wees-etal-2017-dynamic,wang2018dataselection}. This method has four phases: (i) an initial model is pre-trained on
the generic dataset, (ii) this model is fine tuned on the specific data, (iii) the generic set is restricted to the generic data
whose loss improvement between the pre-trained model (i) and the fine tuned model (ii) is the greatest. Finally, (iv) the training of the pre-trained model (i) is resumed on the selected data from stage (iii). As we do for all data selection method, we consider further fine tuning the final CDS model on the specific training set. Although CDS is a generic method applicable to any training objective, it enjoys additional properties when applied to generative models trained to maximize the (conditional) training likelihood. It can both be considered an importance sampling method and an influence function based selection method~\citep{grangier2022tradeoffs}.

We also consider a {\it domain classifier}. In that case, a simple model is pretrained on a binary classification problem to distinguish between generic and specific training examples. The model has the same architecture as the weighting model we use with bilevel methods and it minimizes the binary cross entropy on batches with the same proportion of specific and generic data. This model can estimate the probability that an example belongs to the specific set and is applied to restrict the generic set to the data with the highest estimates. We can train a model on this restricted set and later fine tuning on the specific data.

Closer to our bilevel selection methods, we evaluate learning to re-weight, LTR~\citep{ren2018reweight} and meta-weight net~\citep{shu19metaweight}). Learning to re-weight is similar to the DDS approach we presented in Section~\ref{sec:methods} except it does not maintain a weighting model. Instead, at each step, the model considers a uniform distribution over the generic batch. It then computes the gradient of this flat weighting as free parameters with the outer update, Equation~\ref{eq:outer_problem}.
This single step updates from uniform is then used to reweight the generic loss and update the main model, Equation~\ref{eq:inner_problem}. Compared to our work, this method does not persist a weighting model across steps and does not
allow learning complex distributions. The lack of weighting model is also less efficient since a generic example $x$ cannot be discarded without evaluating the main model and its gradient at $x$. 

Meta-weight net is a particular, simple case of DDS in which the weight model takes as input a single scalar for each example:
the example loss, i.e. $w(x; \alpha) = {\text{mlp}}({\ell}(x; \theta); \alpha)$. This parameterization is sensible for some applications, e.g. loss based up-weighting is a common approach for active learning in classification problems with little intrinsic uncertainty~\citep{settles2009active}. Loss values can be indicative of an example difficulty and loss based up-weighting  might accelerate learning. However, an example loss seems to be orthogonal to the example value for domain transfer.

\subsection{Language Modeling}
\label{subsec:lm}
Our language modeling (LM) experiments relies on two datasets,
the C4 dataset~\citep{raffel2019c4} is used as the generic 
set and the RCV1~\citep{lewis04rcv1} dataset is used as the 
specific set. C4 is a dataset of English language 
web pages from common crawl~\citep{patel2020commoncrawl}, while 
RCV1 consists of newswire stories from Reuters. This setup is 
representative of a generic large corpus spanning different 
types of examples (c4) while the specific task contains an 
homogeneous set of examples from the same domain and from the 
same source (RCV1). In our setup, 
we use 30m examples from C4 and 10k examples from RCV1.

\begin{table}[t]
  \begin{center}
    \caption{Model architectures}
    \label{tbl:arch}
    \begin{tabular}{p{0.95\textwidth}} 
      \textbf{Language Model}\\\hline
      {\it Main model:} Transformer decoder with 12 layers, 8 attention heads, residual dimension of 256, feed-forward latent dimension of 1,024.\\
      {\it Weight model:} Convolutional network with 2 layers followed by mean pooling, latent dimension of 128.\\~\\
      \textbf{Translation Model}\\\hline
      {\it Main model:} Transformer with 6 encoder layers and 6 decoder layers, 16 attention heads, residual dimension of 1,024, feed-forward latent dimension of 4,096.\\
      {\it Weight model:} Embedding layer of dimension 32 followed by an MLP with a latent dimension of 128.\\
      ~\\
      \textbf{Image Classifier}\\\hline
      {\it Main model:} Dual encoder clip model with ResNet 50 for images (224x224) and an multi-layer perceptron (2 latent layers with dim. 768) on top of sentence BERT for text.\\
      {\it Weight model:} Convolutional network over 32x32 images with 4 layers of dimension 32, 32, 32 and 64.\\
    \end{tabular}
  \end{center}
\end{table}

Our language model is a byte-level language model based on the
transformer decoder architecture~\citep{vaswani17transformer}.
Although sub-word language models are more common than 
byte-level ones~\citep{sennrich-etal-2016-neural,al2019character}, 
we rely on
bytes to avoid our out-of-domain generalization 
results to be contaminated by the mismatch between the evaluation 
data and the tokenizer training data~\citep{rust-2021-tokenizer}.
The weighting network is a small convolutional network.
Table~\ref{tbl:arch} gives architectural details.
We also use the same architecture for the domain classifier baseline.
We report performance in terms of log-perplexity, i.e. negative log
likelihood. Our implementation is derived from the language model
of the Flax library~\citep{heek2020flax}.

Table \ref{tbl:lm_main} reports the results of our language modeling experiments.
In general, domain adaptation is beneficial in our setting. The only method trained
exclusively on c4 (baseline without fine-tuning) is much worse than all alternatives
except for MetaWeightNet. Before pretraining, mixing is the only method which directly
applies updates from the specific training data and it performs best. The other methods 
only emphasize part of the generic set without applying specific updates during
pretraining. This emphasis already show a benefit at pretraining time. More importantly,
this benefit is complementary to fine tuning~\citep{iter2021complementarity} 
and these methods yield better results that mixing+fine-tuning. Among them, bilevel
methods perform best, with SOBA giving the highest held-out specific likelihood.

We perform additional language modeling experiments with different domains. We take
9 domains from~\citep{pile} and relies on 10k specific document for each domain. 
The generic set (c4 dataset) and the model architectures are the same as in the previous 
experiments. Data selection methods show that it is helpful to emphasize part of the 
generic set and that this emphasis is complementary to the benefit of fine tuning. The
benefit varies across domains. For instance openweb is similar to the generic set c4 
and only modest gains are observed, while freelaw contains legal proceedings whose domain
is surely relatively rare in c4. Among methods, CDS and classifier provides a strong benefit
for some datasets, but only SOBA consistently ranks among the best methods.

\begin{table}
\begin{minipage}[t]{0.43\columnwidth}
  \begin{center}
    \caption{Language modeling: Log-perplexity (negative log-likelihood per byte) on specific (Reuters). Circled numbers indicate the best results.\label{tbl:lm_main}}
    ~\\
    \begin{tabular}{l | r | r r}
    Method            &     Pre-train &   Fine-tune &\\\hline\hline
    Baseline          &         1.197 &       0.864 &\\
    Mixing            &         0.860 &       0.846 &\\
    CDS               &         1.071 &       0.830 &\\
    Domain classif.   &         1.099 &       0.892 &\\
    MetaWeightNet     &         1.212 &       0.867 &\\
    LTR               &         1.150 &       0.877 &\\
    Sparse DDS        &         1.033 &     \cB  0.822 \\
    Sparse Anograd    &         1.035 &     \cB  0.822 \\ 
    Sparse SOBA       &         1.018 &     \cA  0.819 \\\hline
    \end{tabular}\\
    \end{center}
\end{minipage}\hfill % maximize horizontal separation
\begin{minipage}[t]{0.51\columnwidth}
  \begin{center}
    \caption{Machine translation: BLEU and loss on specific (newstest2020).\label{tbl:mt_main}}
    ~\\~\\
    \begin{tabular}{l | r r | r r}
    Method            & \multicolumn{2}{c|}{Pre-train}  & \multicolumn{2}{c}{Fine-tune}\\
                      & BLEU      & loss    & BLEU & loss \\\hline\hline
                      
    Baseline          &     27.63 &    2.56 &        34.06 &    2.53\\
    Mixing            &     31.34 &    2.60 &        33.11 &    2.69\\
    CDS               &     34.14 &    2.53 &        34.25 &    2.53\\
    Domain classif.   &     35.56 &    2.37 &   \cA  38.03 &    2.35\\
    MetaWeightNet     &     26.81 &    2.59 &        33.34 &    2.53\\
    LTR               &     28.60 &    2.73 &        31.15 &    2.71\\
    Sparse DDS        &     33.53 &    2.46 &        35.83 &    2.44\\
    Sparse Anograd    &     36.06 &    2.41 &   \cB  37.28 &    2.40\\
    Sparse SOBA       &     34.23 &    2.39 &   \cC  37.16 &    2.38\\\hline
    \end{tabular}\\
    \end{center}
\end{minipage}
\end{table}

\begin{comment}
Mixing
0.860	4u9hqmpi3q	300	bolt.name=c4lm downsampling=1.00 lr=0.00200 gen_w=0.900
0.846	af4ku9bvm3	300	bolt.name=c4lm downsampling=1.00 gen_w=0.000 ft280_4u9hqmpi3q_mixing

Soba 
1.018	i3tgyyyz7r	300	bolt.name=c4lm soba downsampling=128.00 lr=0.00200
0.819	2bmwrvrns9	346	bolt.name=c4lm downsampling=1.00 gen_w=0.000 ft280_i3tgyyyz7r_soba

Anograd
1.035	5qcbc2tefm	430	bolt.name=c4lm anograd downsampling=64.00 lr=0.00200
0.822	9a2pe3643w	353	bolt.name=c4lm downsampling=1.00 gen_w=0.000 ft280_5qcbc2tefm_anograd

Soft
1.033	a78b5uqumn	500	bolt.name=c4lm soft downsampling=128.00 lr=0.00200
0.822	qi936um5hz	353	bolt.name=c4lm downsampling=1.00 gen_w=0.000 ft280_a78b5uqumn_soft

CDS
1.071	vkxn8dzgt9	496	bolt.name=c4lm cds lr=0.00200 cds pre=280k ft=100k
0.830	nrw6tzsarw	476	bolt.name=c4lm downsampling=1.00 gen_w=0.000 ft280_vkxn8dzgt9_cds

Classifier
1.099	xx9d5eudb8	482	bolt.name=c4lm classifier downsampling=256.00
0.892	f6jfufih4w	318	bolt.name=c4lm downsampling=1.00 gen_w=0.000 ft280_xx9d5eudb8_classifier

Base
1.197	mq38hzu3wx	484	bolt.name=c4lm downsampling=1.00 lr=0.00200
0.864	phymwp4c6w	290	bolt.name=c4lm downsampling=1.00 lr=0.00200 gen_w=0.000 ft280_mq38hzu3wx_base
\end{comment}

\begin{table}
  \begin{center}
    \caption{Language modeling: Log-perplexity on specific for Pile domains.\label{tbl:lm_pile}}
    ~\\~\\
    \begin{tabular}{l | @{\hspace{0.3\tabcolsep}}r@{\hspace{0.5\tabcolsep}} | @{\hspace{0.3\tabcolsep}}r@{\hspace{0.5\tabcolsep}} | @{\hspace{0.3\tabcolsep}}r@{\hspace{0.5\tabcolsep}} | @{\hspace{0.3\tabcolsep}}r@{\hspace{0.5\tabcolsep}} | @{\hspace{0.3\tabcolsep}}r@{\hspace{0.5\tabcolsep}} | @{\hspace{0.3\tabcolsep}}r@{\hspace{0.5\tabcolsep}} | @{\hspace{0.3\tabcolsep}}r@{\hspace{0.5\tabcolsep}} | @{\hspace{0.3\tabcolsep}}r@{\hspace{0.5\tabcolsep}} | @{\hspace{0.3\tabcolsep}}r@{\hspace{0.5\tabcolsep}}}
Method       &   arxiv   & europarl  & freelaw   &   gutenb. & opensub.  & openweb.  &  pmed abs & stackex.   & wikipedia   \\\hline\hline
Base         &     1.438 &     2.219 &     1.555 &     1.365 &     1.277 &     1.220 &     1.088 &      1.313 &     1.103\\
 + ft.       &     0.898 &     0.993 &     0.603 &     0.488 &     1.058 &     1.180 &     0.870 &      1.039 &     0.877\\\arrayrulecolor{lightgray}\hline\arrayrulecolor{black}
Mixing       &     0.909 &     1.019 &     0.606 &     0.487 &     1.067 &     1.153 &     0.874 &      1.049 &     0.874\\
 + ft.       &     0.899 &     1.081 &     0.600 &     0.478 &     1.059 &     1.156 &     0.860 &      1.042 &     0.865\\\arrayrulecolor{lightgray}\hline\arrayrulecolor{black}  
CDS          &     1.216 &     1.981 &     1.284 &     1.253 &     1.193 &     1.154 &     0.829 &      1.100 &     0.944\\
 + ft.       & \cA 0.861 &     0.977 &     0.614 &     0.482 & \cB 1.039 & \cC 1.131 &     0.791 & \cA  0.971 & \cC 0.823\\\arrayrulecolor{lightgray}\hline\arrayrulecolor{black}  
Classifier   &     1.293 &     1.608 &     1.133 &     1.202 &     1.266 &     1.139 &     0.787 &      1.159 &     0.914\\
 + ft.       &     0.920 & \cB 0.892 & \cC 0.582 &     0.481 &     1.066 & \cA 1.125 & \cA 0.765 &      0.998 & \cB 0.807\\\arrayrulecolor{lightgray}\hline\arrayrulecolor{black}  
S. DDS       &     1.231 &     1.868 &     1.184 &     1.285 &     1.288 &     1.290 &     0.828 &      1.112 &     0.988\\
 + ft.       & \cB 0.867 &     0.948 &     0.580 & \cA 0.477 &     1.104 &     1.262 &     0.792 & \cB  0.977 &     0.839\\\arrayrulecolor{lightgray}\hline\arrayrulecolor{black}  
S.Anograd   &     1.219 &     1.659 &     1.237 &     1.274 &     1.193 &     1.150 &     0.814 &      1.132 &     0.989\\
 + ft.       & \cC 0.871 & \cC 0.895 & \cB 0.580 & \cA 0.477 & \cC 1.042 &     1.133 & \cC 0.784 &      0.988 &     0.838\\\arrayrulecolor{lightgray}\hline\arrayrulecolor{black}  
S. SOBA      &     1.210 &     1.582 &     1.124 &     1.296 &     1.184 &     1.149 &     0.803 &      1.134 &     0.908\\
 + ft.       &     0.872 & \cA 0.883 & \cA 0.579 & \cB 0.480 & \cA 1.035 & \cB 1.128 & \cB 0.779 & \cC  0.989 & \cA 0.803\\\hline  
    \end{tabular}
\end{center}
\end{table}

\subsection{Machine Translation}

Our machine translation (MT) experiments learn a translation model from
English into German. They rely on two datasets: our generic set 
is the Paracrawl dataset (Release 1 for WMT 2018) with 36m 
sentence pairs~\citep{banon2020paracrawl}. Our specific set 
concatenates the WMT newstest sets (2009--2019) with source original English sentences, which amounts to 10,015 sentence pairs~\citep{akhbardeh-etal-2021-findings}. We use the 2020 
newstest data (1,997 sentences) as our validation set and 
leave the 2021 newstest data (1,418 sentences) as our test set. Our generic set is therefore a large crawled set with 
different types of text and varying translation quality while the 
specific set is a small set from a single domain with high quality 
translation.

Our translation system is a sub-word model based on the transformer
encoder-decoder architecture. For the weighting network and the domain classifier we compose a shared embedding layer for source and target and apply a multi-layered perceptron on the 
contatenated averaged embeddings of the source and target sentences.
Table~\ref{tbl:arch} gives architectural details.
Our evaluation relies on BLEU scores~\citep{papineni2002bleu} 
for beam-4 decodes.
We also reports some results in terms of loss (i.e. negative log likelihood with label smoothing strength of $0.1$).  Our implementation is derived from the translation model of the Flax library~\citep{heek2020flax}.

Table \ref{tbl:mt_main} reports the results of our machine translation experiments. In that case, SOBA and Anograd provide a strong improvement over the baseline, both before (more than +7 BLEU) and after fine tuning (more than +3 BLEU). However in this setting, the domain classifier is even more effective.
Both for language modeling and for machine translation, we remark that MetaWeightNet performs poorly. MetaWeightNet predicts the selection weight from the loss on the example. This is a common strategy in active learning for classification~\citep{settles2009active}. This notably assumes that the loss per example is indicative of how well the model perform on them. However, in the case of language modeling and translation, the loss per example also reflects the intrinsic entropy of the examples which varies greatly across examples. It therefore seems difficult to use the loss the sole feature for data selection for these tasks. 
Comparing the results of LTR and DDS is also interesting as the methods are similar. DDS maintains a weighting model across steps, while LTR just adapt the distribution for each batch and does not persist cross-steps selection parameters. The benefit of DDS tells that the stability across steps and the lesser dependency on the batch size are important.

\subsection{Image Classification}

Our vision setup performs contrastive training over image and captions -- CLIP~\citep{radford2021learning} -- for generic training and image classification for specific training. Specifically, 
contrastive learning should select the correct caption within a large set of random captions. This approach also allows to perform classification by representing classes as captions of the form "a photo of a <class name>" and letting the model infer the most appropriate caption within that set. As datasets, we rely on yfcc15m~\citep{radford2021learning} for generic training (14.9m image/caption pairs) and ImageNet67~\citep{eshed2020novelty_detection} dataset for the specific task. Imagenet 67 consists in 67 high 
level classes over Imagenet~\citep{deng2009imagenet}, e.g. building, person, fish, dog... Like for other experiments, we consider a setup 
with limited specific data and take 2,010 specific examples, 30 per class, for training. Held-out evaluation is performed with 50 images per class.

For our CLIP model, the image branch is a Resnet 50~\citep{he2016resnet} while the text branch applies an MLP 
over precomputed sentence embeddings from Sentence BERT~\citep{reimers-2019-sentence-bert}. Training applies 
contrastive learning only over alternative captions: for the generic loss, we consider a buffer of past captions as negatives; for the specific loss, we consider all the other class captions as negatives. Our weighting network is a small convolutional network over low resolution images (32x32). Table~\ref{tbl:arch} gives architectural details.

\begin{table}[t]
  \begin{center}
    \caption{Image Classification: Accuracy on specific (ImageNet67).\label{tbl:vision_main}}
    ~\\~\\
    \begin{tabular}{l | r r| r r}
    Method            & \multicolumn{2}{c|}{Pre-train}     & \multicolumn{2}{c}{Fine-tune} \\
                      &  Acc. & Loss     &  Acc. & Loss \\\hline\hline  
    Baseline          &  41.1 &  2.694  &     54.9 &  1.902 \\
    Mixing            &  55.1 &  1.928  &  \cC   55.1 &  1.928  \\
    CDS               &  42.3 &  2.512  &  \cB   55.2 &  1.957 \\
    Domain classif.   &  44.1 &  2.571  &  \cA   57.5 &  1.949 \\
    MetaWeightNet     &  35.5 &  2.743  &     43.9 &  2.351 \\
    LTR               &  36.4 &  2.712  &     44.9 &  2.364 \\
    Sparse DDS        &  40.5 &  2.609  &     53.2 &  2.067 \\
    Sparse Anograd    &  41.4 &  2.563  &     53.6 &  2.055 \\ 
    Sparse SOBA       &  41.1 &  2.622  &     53.9 &  2.057\\\hline
    \end{tabular}\\
    \end{center}
\end{table}

\begin{comment}
pretr  -> finetu 
-------------------------------------------------------------
44.111 -> 57.536 vijnvnjcdm classifier ds=4 (tiny cnn) ft=zd7k2nky5f
55.078 -> 54.785 rvaz7jvdb7 mixing gw=0.9950 (ft=br3ggd23xf)
41.106 -> 54.904 mzjzdtzkr4 baseline (generic train only, ft=tm5yae2vz5) 
41.436 -> 53.559 bnun4cj5ga anograd meta_model=tiny_cnn ds=8  (ft = qhjs6xjx4h)
41.106 -> 53.559 gvsyn9ecf8 soba (ft=3ushpfvw4d)
40.535 -> 53.200 g78aesu5gy soft (ft=nz8zsnbjj6)
42.278 -> 55.233 zsxwjirtks cds  (ft=pcwzhdd5qj)
36.418 -> 44.916 igz4cwg38n LTR (ft=c45u3kd3qm)
35.517 -> 43.899 e4k5ujefc9 MetaWeightNet (ft=j553pqfsdn)
[41.466 -> 54.515 4p9kn99ffh baseline (generic train only, ft=r6keddmp84 ) deprecated]
\end{comment}

Table \ref{tbl:vision_main} reports the results of our image classification experiments. Unlike for our text experiments, 
the benefit of data selection is limited for this task. After fine-tuning, only the CDS and domain classifier methods outperform 
the baseline model. The bilevel data selection methods do not outperform the baseline method. We shed light on the cause of this 
poor performance in our further analysis in Section~\ref{sec:grad_alignment_discriminative_power}.

\section{Analysis}

\subsection{Learning a Distribution vs Learning a Curriculum}
\label{sec:curriculum}

Algorithm~\ref{alg:online_bilevel_selection} produces a sequence of main model's parameters $\theta_t$ and weighting model's parameters $\alpha_t$, that go towards the solution of the bilevel problem (\ref{eq:outer_problem}).
We investigate whether the weighting model's parameters correspond to a \emph{curriculum}:
does the evolution of the weighting parameters $\alpha_t$ adapt to the particular data
needed at each step, helping the model perform better than a fixed weighting scheme?

We are in the LM task setup described in Section~\ref{subsec:lm}, except that we use a smaller ``large batch size'' $B^{\text{big}}_{\text{generic}}$ (see Section~\ref{subsec:big_batch_trick}). We run Algorithm~\ref{alg:online_bilevel_selection} with SOBA to obtain a sequence $\theta_t, \alpha_t$. We then compare this setting with two new training runs with standard ERM using different data weighting:
\\
-{\it Final weighting:} a new main model is trained with fixed weighting from the weighting model $\alpha_T$.
\\
-{\it Shuffled weighting:} a new main model is trained with a random permutation $\sigma$ of the weights $\alpha_{\sigma(t)}$.
\\
Table~\ref{tbl:curriculum} shows that SOBA's curriculum is not beneficial compared
to the fixed final weighting scheme on this task. The lesser performance of shuffled weighting certainly
highlight poor weighting from early $\alpha_t$.
Note that the results reported in this section do not match Section~\ref{subsec:lm} because of a smaller $B^{\text{big}}_{\text{generic}}$ was used in this ablation.
\begin{table}
\begin{minipage}[t]{0.47\columnwidth}
  \begin{center}
    \caption{Does the weight model's trajectory correspond to a curriculum? Pre-train LM log perplexity on Reuters for different trajectories.\label{tbl:curriculum}}~\\
    \begin{tabular}{l|r}
    Weight model trajectory & log-perplexity \\  \hline\hline
    SOBA curriculum  & 1.047 \\
    Final weighting         & 1.044 \\
    Shuffled weighting      & 1.055 \\\hline
    \end{tabular}
  \end{center}
\end{minipage}\hfill % maximize horizontal separation
\begin{minipage}[t]{0.47\columnwidth}
  \begin{center}
    \caption{Alternative sampling strategies to filter the generic batch. Pre-train LM log perplexity on Reuters.\label{tbl:sampling}}~\\
    \begin{tabular}{l|r}
    Sampling strategy              & log-perplexity \\  \hline\hline
    Importance Sampling            & 2.038 \\
    Sampling without replacement   & 1.040 \\
    Selecting the highest weights  & 1.227 \\\hline
    \end{tabular}
  \end{center}
\end{minipage}
\end{table}

\begin{comment}
Trajectory experiments:
Base run     as7py4hgw5
Fixed        7gmxu7jfqa 
Random perm. ewa5yzfui6

Sampling experiments:
Importance Sampling            2whx44pnzd
Sampling without replacement   a2enaxi9sa 
Selecting the highest weights  56wfjss54s
\end{comment}

\subsection{Big Batches: Importance Sampling vs Filtering}
\label{sec:sampling}

In Algorithm~\ref{alg:online_bilevel_selection}, we denote ${\rm filter}(B, \alpha,  n)$ the operation resulting in a smaller sub-batch of size $n$ starting from the generic batch $B$ using the weighting network parameterized by $\alpha$. To get an unbiased estimate of the re-weighted generic loss, one can apply {\it importance sampling} and sample (with replacement) from the weight distribution induced by $\alpha$ on B. Alternatively one can instead sample {\it without replacement} from that distribution or restrict the batch B to its {\it highest weighted} elements. The last two alternative are biased. Nevertheless, our results in Section~\ref{sec:experiments} uses sampling without replacement. 

Table~\ref{tbl:sampling} justifies this choice. Basically, we observe that the learned weighted distribution is concentrated along few examples which yield importance sampling batches to contain less diverse sets than when sampling with replacement. Similarly, cutting the tail of the distribution (highest weights selection) drop lower weighted -- but still helpful -- examples. These experiments illustrate that gradient-based estimates fail to account for the long term benefit of a more diverse training set. Although sampling with replacement alleviates this issue, more principled solutions should be investigated in future work.

\subsection{On the Discriminative Power of Gradient Aligments}
\label{sec:grad_alignment_discriminative_power}

Our experiments highlight that bilevel optimization for data selection performs differently across tasks. 
We explore if a simple diagnostic could help understand these differences. Our method considers a base 
model $\theta_t$ trained on the generic distribution for $t$ steps. We take a diagnostic batch 
$B_{\text{mix}}$ which blends unseen generic and specific data in equal proportion. We want to verify 
how the weighting model on the mixed data would move away from a uniform weighting scheme in an outer update.
We want to observe whether the weighting model would increase the weights of specific examples if some of 
these were "hidden" within the generic set.

For DDS and Anograd, increase or decrease in weights depends on the alignment 
between individual example gradients from $x \in B_{\text{mix}}$ and 
the expected gradient on the training specific batch $B_{\text{specific}}$.
$$
a(x, B_{\text{specific}}) 
= \nabla_\theta \ell(x, \theta)^\top ~ \nabla_\theta \ell(B_{\text{specific}}, \theta) 
$$
between individual example gradients from $x \in B_{\text{mix}}$ and 
the expected gradient on the training specific batch $B_{\text{specific}}$,
denoted as
$
\ell(B_{\text{specific}}, \theta) 
:= \frac{1}{|B_{\text{specific}}|} \sum_{x\in B_{\text{specific}}}  \ell(x, \theta).
$
We then normalize the batch gradient and define,
$$
a_{\text{norm}}(x, B_{\text{specific}}) 
= \nabla_\theta \ell(x, \theta)^\top ~
\frac{\nabla_\theta \ell(B_{\text{specific}}, \theta)}{\|\nabla_\theta \ell(B_{\text{specific}}, \theta)\|}.
$$
This normalization allows to take an example $x$ and verify whether its gradient
aligns better with the specific batch gradient than with the generic batch 
gradient, i.e.
$$
a_{\text{norm}}(x, B_{\text{specific}}) > a_{\text{norm}}(x, B_{\text{generic}}).
$$
We report the rate at which this inequality is true for specific 
examples,
$$
\text{SAR} = \mathop{{}\mathbb{E}}_{
\substack{x \sim {\cal D}_\text{specific}\\ 
B_{\text{specific}} \sim \text{Batch}({\cal D}_\text{specific}) \\
B_{\text{generic}} \sim \text{Batch}({\cal D}_\text{generic})}} 
\mathbbm{1}\left\{a_{\text{norm}}(x, B_{\text{specific}}) > a_{\text{norm}}(x, B_{\text{generic}})\right\}
$$
We call this measure the Specific Acceleration Rate, SAR. We would like this rate to be high, 
meaning that, according to the Taylor approximation of the loss, updates collected from a 
batch of specific examples should improve the loss on a given specific examples faster 
than updates collected from a generic batch. Symmetrically, we define the Generic Acceleration Rate, 
$$
\text{GAR} = \mathop{{}\mathbb{E}}_{
\substack{x \sim {\cal D}_\text{generic}\\ 
B_{\text{specific}} \sim \text{Batch}({\cal D}_\text{specific}) \\
B_{\text{generic}} \sim \text{Batch}({\cal D}_\text{generic})}} 
\mathbbm{1}\left\{a_{\text{norm}}(x, B_{\text{generic}}) > a_{\text{norm}}(x, B_{\text{specific}})\right\}
$$
It is also desirable that this rate is high. When SAR, GAR are close to chance (50\%), there are
two possible explanations, (i) either generic and specific batches have the same effect on the model, 
meaning that data selection is unlikely to be helpful since training on generic is already as good as 
training on specific for the purpose of minimizing the specific loss; (ii) alternatively, the linear 
approximation (order 1 Taylor expansion) does not help discriminating between the effect of generic 
and specific examples on the specific loss. In that later case, such a learning problem will be
a challenge for bilevel optimization methods where gradient alignments indicates which part of the 
dataset to upweight.

\begin{table}[]
    \centering
    \caption{Specific \& Generic Acceleration Rates }
    \label{tbl:sar}    
    \begin{tabular}{l l|c|c}
    Task &            &  SAR & GAR \\  \hline\hline
    Language modeling     & (c4, reuters)          & 86.2\% & 69.4\% \\
    Machine translation & (paracrawl, newstest)    & 78.1\% & 68.2\% \\
    Image classification & (yfcc15m, imagenet67)   & 50.3\% & 49.8\% \\\hline
    \end{tabular}
\end{table}

Table~\ref{tbl:sar} reports SAR and GAR for our results. These results are indicative of the empirical 
benefit of bilevel optimization methods for data selection. Language modeling where DDS, Anograd and
SOBA are advantageous, has the highest SAR. Conversely, our image classification problem shows near
random SAR, GAR in line with the poor performance on bilevel methods on this problem. We therefore
consider that measuring SAR/GAR can be a simple but informative diagnostic to assess the potential 
benefit of bilevel methods on a new problem.

\subsection{Re-using Weighting Strategies with Larger Scale Models}
%On the Opportunity of re-Using Weighting Models for Scaling Models}

The weighting network is trained by solving the bilevel problem (\ref{eq:outer_problem}), where the loss function $\ell$ depends on the model's architecture.
We investigate whether a weighting network learned with a small model can be re-used out-of-the-box to train a large model and get good performances on the specific set.
The weighting network is frozen: the large model is trained by solving $\min_{\theta}\sum_{x\in\mathcal{D}_{\mathrm{generic}}}w(x;\alpha)\ell(x;\theta)$ with $\alpha$ fixed to the final parameters of the weighting network trained with the small model.
We perform this experiment on the language modeling task (Section~\ref{subsec:lm}), where the small model and large model architectures are specified in Table~\ref{tbl:scaling_archs}; the large model's architecture is the same as in Section~\ref{subsec:lm}, and it has about ten times more parameters.
We observe that the weighting network learned with the small model transfers to the large architecture and leads to a large decrease in the loss on the specific set, which is only slightly worse than using the Sparse SOBA method on the large model itself.
This means that the weighting network learned at a small scale can then seamlessly be used at a larger scale and lead to significant performance improvement on the specific set.

\begin{table}
\begin{minipage}[t]{0.43\columnwidth}
  \begin{center}
    \caption{Small and base model architectures for scaling up the language modeling task.\label{tbl:scaling_archs}}
    ~\\~\\
    \begin{tabular}{l | c c c | r}
    Model           & Layers & Res. dim. & ff dim. & \# Params \\\hline\hline
    Small           &   4    &  128          & 512               & 824.064 \\
    Large            &  12    &  256          & 1024              & 9.530.880  \\\hline
    \end{tabular}\\
    \end{center}
\end{minipage}\hfill % maximize horizontal separation
\begin{minipage}[t]{0.51\columnwidth}
  \begin{center}
    \caption{Results of the scaling experiment\label{tbl:scaling_results}}
    ~\\~\\
    \begin{tabular}{l | l | r r}
    Model  &   Method                      &   Pre-train &\\\hline\hline
    Small  &   Sparse SOBA                 &       1.292 &\\  %69jveabjhx
    Large   &   Baseline                    &       1.197 &\\  %8k4thuaxpr
    Large   &   Sparse SOBA                 &       1.018 &\\\hline  %ak95b3iu4y
    Large   &   Weights from Small          &       1.034 &\\  %7sivsyzndj
    \end{tabular}\\
    \end{center}
\end{minipage}
\end{table}

\section{Conclusions}

This work studies bilevel optimization for learning training distributions. We consider the setup where a model is trained from two training sets, a large generic set and a small specific set, where only the later is representative of the test conditions. We propose a scalable algorithm that learns a training distribution over the generic data such that the loss on the specific set is minimized. 
We showed that our formulation gathers independently-proposed gradient-based methods for data selection under a common framework. We introduced an algorithm that enables streaming through the generic dataset quickly by examining most of the generic samples with only an inexpensive small auxiliary model. This work reported a comprehensive and realistic empirical comparison of data selection strategies across language modeling, machine translation and computer vision. We studied the conditions in which gradient-based data selection is effective and propose a diagnostic based on gradient alignment to efficiently assess these conditions.

Our work also delineates interesting questions for future work. Conceptually, we observe that gradient-based selection methods fail to reward properly the diversity of the selected samples (Section~\ref{sec:sampling}), which deserves further theoretical study. Empirically, the complementarity between fine-tuning and generic data selection highlights that the updates collected from re-weighted generic training and from specific training are different. The existence of complementary updates and their exploitation might also be possible even when one is presented with a single monolithic training distribution. 

\bibliography{main}
\bibliographystyle{tmlr}

\appendix
\section{Common Settings in Transfer Learning}
\label{sec:transfer_learning}

The transfer learning literature defines various settings to leverage training data from 
a different task and/or distribution. Although not all papers use the same definitions,
Table~\ref{tbl:transfer} presents the most common settings with reference to the 
literature supporting these definitions.
\begin{table}[h]
    \centering
    \caption{Classical Transfer Learning Settings\label{tbl:transfer}}~\\
    \begin{minipage}{0.9\textwidth}
    \hrulefill
    \begin{description}
    \item[Transfer Learning] leverages a source distribution in order to perform better on a target distribution~\citep{thrun1998learning}.
    \item[Multitask Learning] improves generalization by leveraging the information contained in the training signals of related tasks~\citep{caruana1993multitask,caruana1997multitask}.
    \item[Domain Adaptation] aims to improve accuracy on target distribution with insufficient labeled data by leveraging a model trained on a different but related source distribution~\citep{farahani2021brief}.
    \item[Unsupervised Domain Adaptation] considers the setting where labeled source domain data (x, y) are available for training, while only unlabeled (x) data from the target domain are available~\citep{ganin15_uda}.
    \item[Distribution Shift] considers that the test distribution is different from the training distribution, usually in the context where the model cannot be retrained to adapt to the new test conditions~\citep{koh21wilds}.
    \item[Gradual Distribution Shift] is an online setting where the training distribution progressively evolves~\citep{kumar2020understanding}.
    \item[Covariate shift] corresponds to a predictive setting where the distribution over the input features p(x) is different at training and test time, while the posterior distribution p(y|x) does not change~\citep{bickel2009covshift}.
    \item[Label shift] corresponds to a predictive setting where the class prior p(y) between train and test changes but the conditional distribution p(y|x) is assumed identical~\citep{garg2020labelshift}.
    \item[Fine-tuning] is a specific domain adaptation technique which considers
    training a model on the target domain from an initial model trained on the source domain~\citep{matic1993writer,denevi2018learning,zhang2021bertft}.
    \item[Zero-Shot Task Transfer] addresses new tasks at test time without updating the model~\citep{larochelle2008zero}. It typically relies on a way to represent novel tasks in order to condition the model, e.g. text prompts~\citep{radford2019language,srivastava2022beyond}.
    \item[Few-Shot Task Transfer] is similar to zero-shot transfer and does not update the model weights. As a difference, the task conditioning information provides few training instances along with the description of the tasks~\citep{radford2019language,srivastava2022beyond}.
    \end{description}
    \hrulefill
    \end{minipage}
\end{table}
\end{document}